\documentclass[10pt,twocolumn,letterpaper]{article}

\usepackage{cvpr}
\usepackage{times}
\usepackage{epsfig}
\usepackage{graphicx}
\usepackage{amsmath}
\usepackage{amssymb}

\usepackage[breaklinks=true,bookmarks=false]{hyperref}

\cvprfinalcopy 

\def\cvprPaperID{2027} 

\begin{document}

\title{Pattern-Affinitive Propagation across \\Depth, Surface Normal and Semantic Segmentation}
\author{Zhenyu Zhang$^1$$^\dag$ \hspace{0.5cm}
Zhen Cui$^{1}$$^\dag$\thanks{Corresponding authors} \hspace{0.5cm}
Chunyan Xu$^1$\thanks{Zhenyu Zhang, Zhen Cui, Chunyan Xu, Yan Yan and Jian Yang are with PCA Lab, Key Lab of Intelligent Perception and Systems for High-Dimensional Information of Ministry of Education, and Jiangsu Key Lab of Image and Video Understanding for Social Security, School of Computer Science and Engineering, Nanjing University of Science and Technology. Zhenyu Zhang is also a visiting student in University of Trento.}\\
\hspace{0.4cm} Yan Yan$^1$$^\dag$ \hspace{1cm}
Nicu Sebe$^2$\thanks{Nicu Sebe is the head of Dept. of Information Engineering and Computer Science
Leader of Multimedia and Human Understanding Group (MHUG)
University of Trento.} \hspace{0.7cm}
Jian Yang$^{1}$$^{\dag*}$\\
\and
$^1$PCA Lab, Nanjing University of Science and Technology\\
{\tt\small zhangjesse, zhen.cui, cyx, yyan, csjyang@njust.edu.cn}\\
\and
$^2$Multimedia and Human Understanding Group,
University of Trento\\
{\tt\small niculae.sebe@unitn.it}
}

\maketitle

\begin{abstract}
   In this paper, we propose a novel Pattern-Affinitive Propagation (PAP) framework to jointly predict depth, surface normal and semantic segmentation. The motivation behind it comes from the statistic observation that pattern-affinitive pairs recur much frequently across different tasks as well as within a task. Thus, we can conduct two types of propagations, cross-task propagation and task-specific propagation, to adaptively diffuse those similar patterns. The former integrates cross-task affinity patterns to adapt to each task therein through the calculation on non-local relationships. Next the latter performs an iterative diffusion in the feature space so that the cross-task affinity patterns can be widely-spread within the task.
   Accordingly, the learning of each task can be regularized and boosted by the complementary task-level affinities. Extensive experiments demonstrate the effectiveness and the superiority of our method on the joint three tasks. Meanwhile, we achieve the state-of-the-art or competitive results on the three related datasets, NYUD-v2, SUN-RGBD and KITTI.
\end{abstract}

\section{Introduction}\label{sec1}

The predictions of depth, surface normal and semantic segmentation are important and challenging for scene understanding. Also, they have many potential industrial applications such as autonomous driving system~\cite{chen2015deepdriving}, simultaneous localization and mapping (SLAM)~\cite{tateno2017cnn} and socially interactive robotics~\cite{fong2003survey}. Currently, most methods~\cite{eigen2015predicting, eigen2014depth,fouhey2013data, fouhey2014unfolding,Long2017Fully, Noh2015Learning} focused on one of the three tasks, and they also achieved the state-of-the-art performance through the technique of deep learning.
\begin{figure}[!t]
  \centering
\includegraphics[width=7cm]{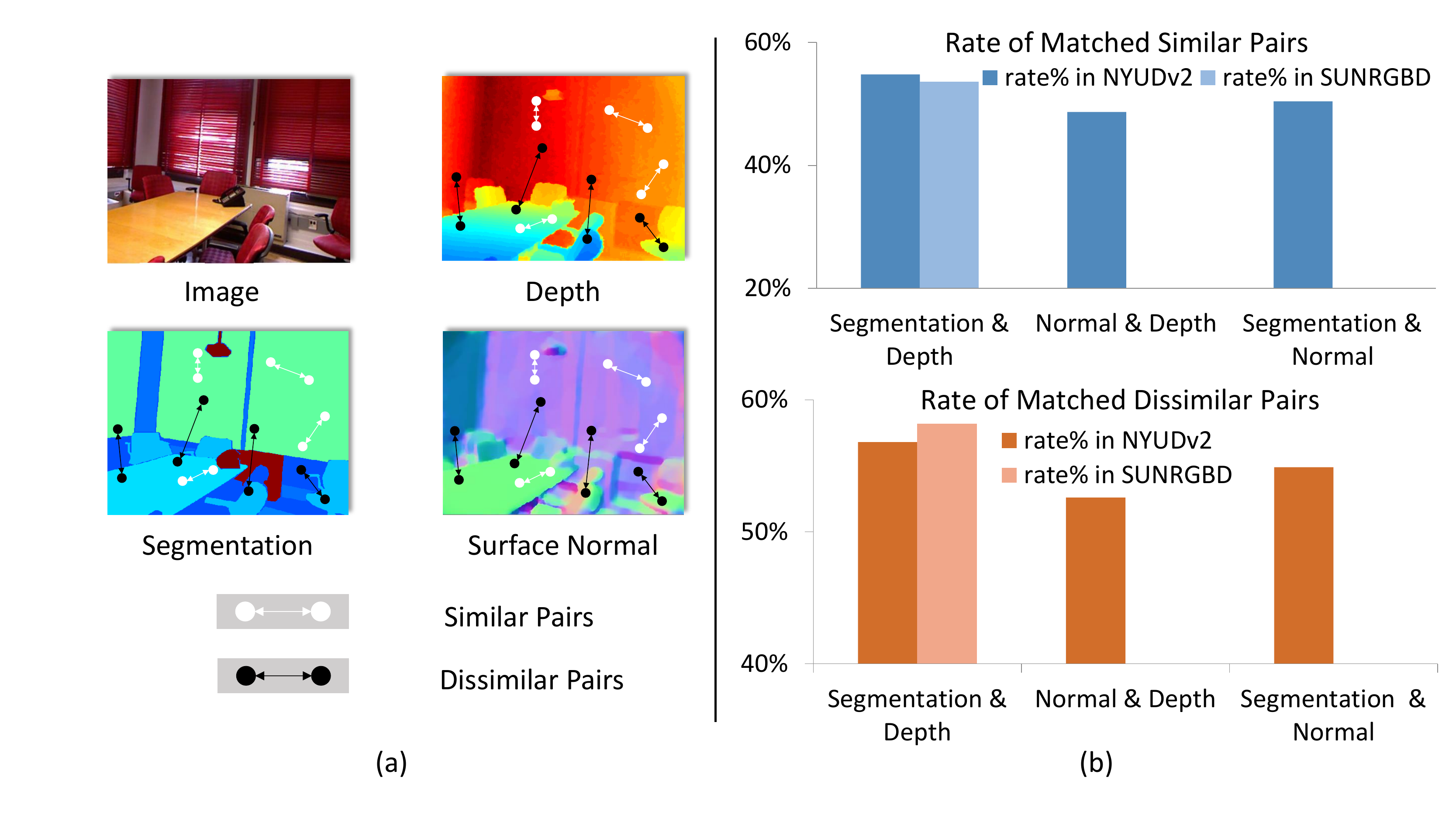}
   \caption{Statistics of matched affinity (or dissimilar) pairs across depth, surface normal and segmentation maps. (a) Visual exhibition. The point pairs colored white are the matched affinity pixels across three tasks at the same positions, while the pairs of black points correspond to dissimilar pixels across three maps. For the similarity metrics, REL/RMSE/Label consistency are taken respectively for the three maps.
   (b) Statistical results. We compute the success ratio of pairs matching across different maps on NYUD-v2 and SUN-RGBD datasets, and observe that the success ratios of pairs matching cross tasks are rather high.
   }\label{fig1}
  \vspace{-0.5cm}
\end{figure}

In contrast to the single-task methods, recently, several joint-task learning methods \cite{xu2018pad, zhang2018joint, qi2018geonet, li2015depth} on these tasks have shown a promising direction to improve the predictions by utilizing task-correlative information to boost for each other. In a broad sense, the problem of joint-task learning has been widely studied in the past few decades~\cite{Caruana1997Multitask}. But more recently most approaches took the technique line of deep learning for possible different tasks~\cite{Misra2016Cross, Girshick2015Fast, He2017Mask, Kim2016Unified, Kokkinos2017UberNet}. However,
most methods aimed to perform feature fusion or parameter sharing for task interaction. The fusion or sharing ways may utilize the correlative information between tasks, but there exist some drawbacks. For examples, the integration of different features might result into the ambiguity of information; the fusion does not explicitly model the task-level interaction where we do not know what information are transmitted. Conversely, could we find some explicitly common patterns across different tasks for the joint-task learning?

We take the three relative tasks: depth estimation, surface normal prediction and semantic segmentation, and then conduct a statistical analysis on those second-order patterns across different tasks on NYUD-v2~\cite{silberman2012indoor} and SUN-RGBD~\cite{Song2015SUN} dataset. First, we define the metric of any two pixels in the predicted images. The average relative error (REL) is used for depth images, the root mean square error (RMSE) is used for surface normal images, and the label consistency is for segmentation images. A pair of pixels have an affinity (or similar) relationship when their error is less than a specified threshold, otherwise they have a dissimilar relationship. Next, we accumulate the matching number of those similar pairs (or dissimilar pairs) with the same space positions across the three types of corresponding images. As shown in Fig.~\ref{fig1}(a), the affinity pairs (colored white points) at the common positions may exist in different tasks. Meantime, there exist some common dissimilar pairs (colored black points) across tasks. The statistical results are shown in Fig.~\ref{fig1}(b), where REL threshold of depth is set to 20\%, and RMSE threshold of surface normal is set to 26\% according to the performances of some state-of-the-art works~\cite{qi2018geonet, bansal2016marr, Laina2016Deeper}. We can observe that the success ratios of matching pairs across two tasks are rather high, and around 50\% - 60\% similar pairs are matched. Moreover, we have the same observation on the matching dissimilar pairs, where REL threshold of depth is set to 20\%, and RMSR threshold of surface normal is set to 40\%.
Anyhow, this observation of the second-order affinities is great important to bridge two tasks.

Just motivated by the statistical observation, in this paper we propose a Pattern-Affinitive Propagation (PAP) framework to utilize the cross-task affinity patterns to jointly estimate depth, surface normal and semantic segmentation. In order to encode long-distance correlations, the PAP utilizes non-local similarities within each task, different from the literatures~\cite{liu2017learning, cheng2018depth} only considering local neighbor relationships. These pair-wise similarities are formulated as an affinity matrix to encode the pattern relationships of the task. To spread the affinity relationships, we take two propagation stages, cross-task propagation and task-specific propagation. The affinity relationships across tasks are first aggregated and optimized to adapt to each specific task by calculating on three affinity matrices. We then conduct an iterative task-specific diffusion on each task by leveraging the optimized affinity information from the corresponding other two tasks. The diffusion process is performed in the feature space so that the affinity information of other tasks can be widely spread into the current task. Finally, the learning of affinitive patterns and the two-stage propagations are encapsuled into an end-to-end network to boost the prediction process of each task.

In summary, our contributions are in three aspects: i) Motivated by an observation that pattern-affinitive pairs recur much frequently across different tasks, we propose a novel Pattern-affinitive Propagation (PAP) method to utilize the matched non-local affinity information across tasks. ii) Two-stage affinity propagations are designed to perform cross-task and task-specific learning. An adaptive ensemble network module is designed for the former while the strategy of graph diffusion is used for the latter. iii) We make extensive experiments to validate the effectiveness of PAP method and its modules therein, and achieve the competitive or superior performances on depth estimation, surface normal prediction and semantic segmentation on NYUD-v2~\cite{silberman2012indoor}, SUN-RGBD~\cite{Song2015SUN}, and KITTI~\cite{uhrig2017sparsity} datasets.

\begin{figure*}[!t]
\centering
\includegraphics[width=15cm]{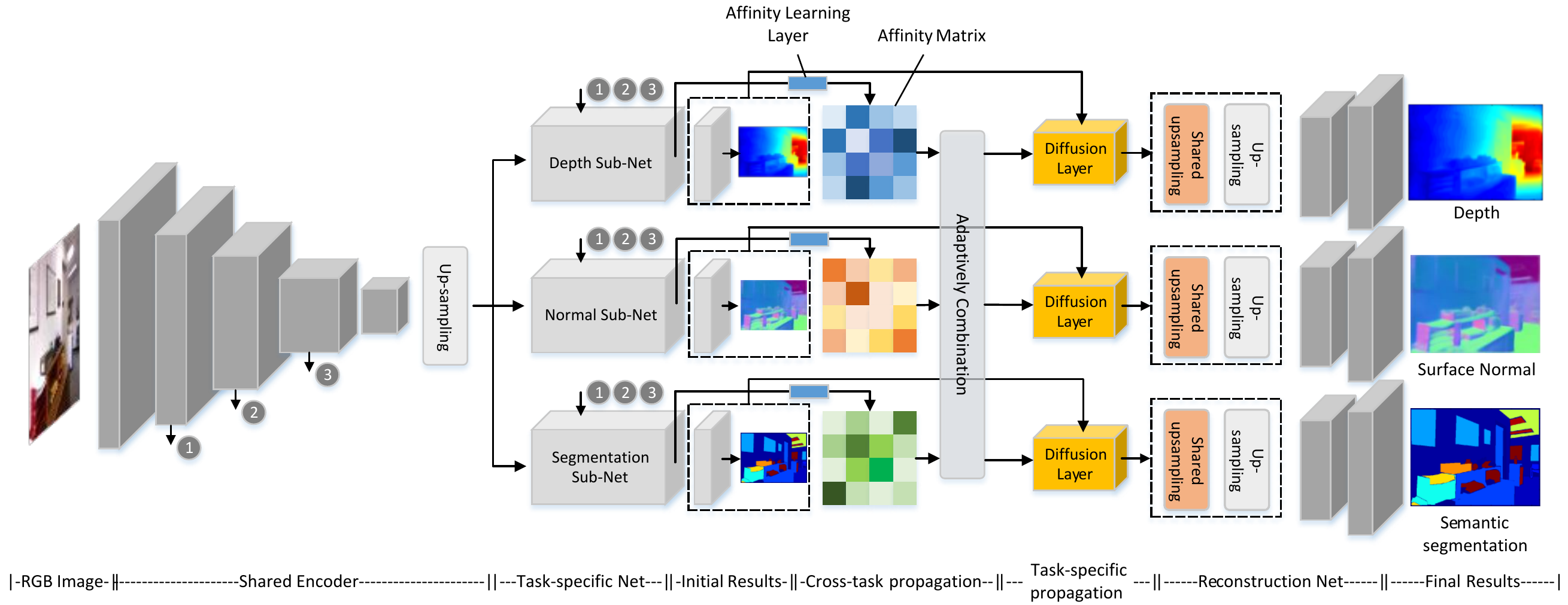}
\centering
\caption{The overview of our Pattern-Affinitive Propagation network for jointly predicting depth, surface normal and semantic segmentation. The initial predictions are produced from each task-specific network. During cross-task propagation, the network firstly learns an affinity matrix by affinity learning layer to represent the pair-wise relationships of each task, then adaptively combines these matrices to propagate the cross-task affinitive patterns.
Note that, the combined affinity matrices is different for each task. Then we use the combined matrix to conduct task-specific propagation by a diffusion layer, propagating the affinitive patterns back to the features for each task.
Finally the diffused features are applied to three reconstruction networks to produce the final results with higher resolution.}
\label{pic2} \vspace{-0.5cm}
\end{figure*}
\section{Related Works}

\textbf{Depth Estimation:} Many works have been proposed for monocular depth estimation~\cite{eigen2015predicting, eigen2014depth, liu2015Learning, li2015depth, Kundu_2018_CVPR, Laina2016Deeper, zhou2017unsupervised, Wang2015Towards, Roy2016CVPR, xu2018structured, xu2018pad, qi2018geonet, zhang2018joint}. Recently, Xu~\textit{et al.}~\cite{Xu2017CVPR} employed multi-scale continuous CRFs as a deep sequential network for depth prediction. Fu \textit{et al.}~\cite{fu2018deep} tried to consider the ordinal information in depth maps and designed a ordinal regression loss function. 

\textbf{RGBD Semantic Segmentation:} As the large RGBD dataset was released, some approaches~\cite{Gupta2014Learning, he2017std2p, SJ2017RDF, Deng2015Semantic, He2016STD2P, Li2016LSTM} attempted to fuse depth information for better segmentation. Recently, Qi \textit{et al.} \cite{qi20173d} designed a 3D graph neural network to fuse the depth information for segmentation. Cheng \textit{et al.} \cite{Cheng2017Locality} computed the important locations from RGB images and depth maps for upsampling and pooling. 

\textbf{Surface Normal Estimation:} Recent methods designed for surface normal estimation are mainly based on deep neural networks~\cite{fouhey2013data, fouhey2014unfolding, zeisl2014discriminatively, wang2016surge}. Wang \textit{et al.} \cite{wang2015designing} designed a network to incorporate local, global and vanishing point information for surface normal prediction. In work of \cite{bansal2016marr}, a skip-connected architecture was proposed to fuse features from different layers for surface normal estimation. 3D geometric information was also utilized in \cite{qi2018geonet} to predict depth and normal maps. 

\textbf{Affinity Learning:} Many affinity learning methods were designed based on physical nature of the problems \cite{he2013guided, krahenbuhl2011efficient, levin2008closed}. Liu \textit{et al.} \cite{liu2016learning} improve the modeling of pair-wise relationships by incorporating many priors into diffusion process. Recently, work of \cite{bertasius2017convolutional} proposed an convolutional random walk approach to learn the image affinity by supervision. Wang \textit{et al.} \cite{wang2017non} proposed a non-local neural network to mine the relationships with long distances.
Some other works \cite{liu2017learning, cheng2018depth, Ke_2018_ECCV} tried to learn local pixel-wise affinity for semantic segmentation or depth completion. Our method is different from these approaches in the following aspects: needs no prior knowledge and is data-driven; needs no task-specific supervisons; learns the non-local affinity rather than limited local pair-wise relationships; learns the cross-task affinity information rather than learning the single-task affinity for task-level interaction.

\section{Non-Local Affinities}\label{tal}

Our aim is to model the affinitive patterns among tasks, and utilize such complementary information to boost and regularize the prediction process of each task. According to our analysis aforementioned, we want to learn the pair-wise similarities and then propagate the affinity information into each task. Instead of learning local affinities as literature \cite{liu2017learning, cheng2018depth}, we attempt to utilize non-local affinities, which also recur frequently as illustrated in Fig.~\ref{fig1}. Formally, suppose $\mathbf{x}_i,\mathbf{x}_j$ are the feature vectors of the $i$-th and $j$-th positions, we can define their similarity $s(\mathbf{x}_i,\mathbf{x}_j)$ through some functions such as L1 distance $\|\textrm{x}_{i} - \textrm{x}_{j}\|$, inner product $\mathbf{x}_i^T\mathbf{x}_j$, and so on. We employ the exponential function ($e^{s(\cdot,\cdot)}$ or $e^{-s(\cdot,\cdot)}$) to make the similarities non-negative and larger for those similar pairs than dissimilar pairs. To reduce the influence of scale, we normalize the similarity matrix $\mathbf{M}$ into $\mathbf{M}_{ij}/\sum_k \mathbf{M}_{ik}$, where $\mathbf{M}$ is the matrix of pair-wise similarities across all pixel positions. In these ways, the matrix $\mathbf{M}$ is \textbf{symmetric}, has \textbf{non-negative} elements and \textbf{finite Frobenius norm}.
Accordingly, for the three tasks, we can compute their similarity matrices $\mathbf{M}_\textrm{depth}, \mathbf{M}_\textrm{seg}, \mathbf{M}_\textrm{normal}$ respectively. According to the above statistic analysis, we can propagate the affinities by integrating the three similarity matrices for one specific task, which will be introduced in the following section.

\section{Pattern-Affinitive Propagation}
In this section, we introduce the proposed Pattern-Affinitive Propagation (PAP) method. We efficiently implement the PAP method into a deep neural network through designing a series of network modules. The details are introduced in the following.

\subsection{The Network Architecture}

We implement the proposed method into a deep network as shown in Fig.~\ref{pic2}, which depicts the network architecture. The RGB image is firstly fed into a shared encoder (e.g., ResNet \cite{he2016deep}) to generate hierarchical features. Then we upsample the features of the last convolutional layer and feed them to three task-specific networks. Note that we also integrate multi-scale features derived from different layers of encoder with each task-specific network, as shown by the gray dots. Each task-specific network has two residual blocks, and produces the initial prediction after a convolutional layer. Then we conduct cross-task propagations to learn the task-level affinitive patterns. Each task-specific network firstly learns an affinity matrix by the affinity learning layer to capture the pair-wise similarities for each task, and secondly adaptively combine the matrix with other two affinity matrices to integrate the task-correlative information. Note that, the adaptively combined matrix is different for each task. After that, we conduct task-specific propagation via a diffusion layer to spread the learned affinitive patterns back to the feature space.
In each diffusion process, we diffuse both initial prediction and the last features from each task-specific network by the combined affinity matrix.

Finally, the diffused features of each task are fed into a reconstruction network to produce final prediction with higher resolution. We firstly use a shared and a task-specific upsampling block to upscale the feature maps. Each upsampling block is built as a up-projection block \cite{Laina2016Deeper}, and parameters in the shared upsampling block are shared for every task to capture correlative local details. After the upsampling with the two blocks, the features are concatenated and fed into a residual block to produce final predictions.
The scale factor of each upsampling block is set to 2, and the final predictions are half of the input scale. This means that the number of upsampling blocks depends on the scale on which we want to learn affinity matrix. In experiments, we learn affinity matrices on 1/16, 1/8 and 1/4 input scale, which means there are 3, 2 and 1 upsampling stages in the reconstruction network respectively.
The whole network can be trained in an end-to-end manner, and the details of the cross-task and task-specific propagations will be introduced in the following sections.

\subsection{Cross-Task Propagation}
\begin{figure}[!t]
  \centering
\includegraphics[width=7cm]{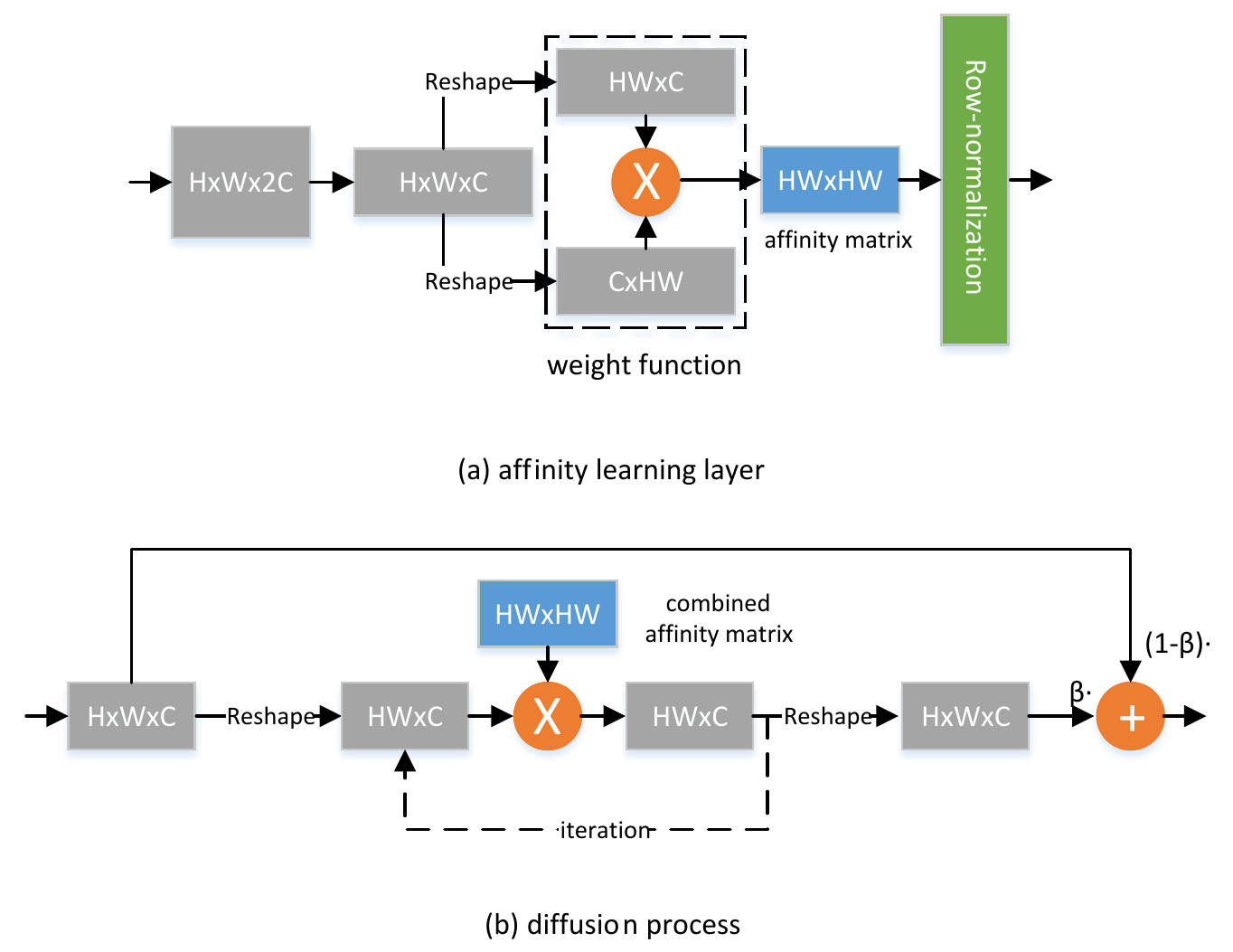}
   \caption{The detailed information of affinity learning layer and diffusion process, and each block describes the feature and its shape. $\otimes$ represents the matrix multiplication.
   (a) affinity learning layer. The dashed box is corresponding to the function for computing similarities, and we only illustrate the dot-product as an example. (b) diffusion process. $\oplus$ represents the weighted sum with a parameter $\beta$.
   The dashed arrows are only performed when the iteration is not finished.
   }\label{fig3}
  \vspace{-0.5cm}
\end{figure}
In this section we elaborate how to conduct cross-task propagation. Firstly, we learn an affinity matrix by affinity learning layer to represent the pair-wise similarities for each task.
The detailed architecture of the affinity learning layer can be observed in Fig.~\ref{fig3}(a). Assuming the feature generated by the last layer of each task-specific network is $\textrm{F}\in \mathbb{R}^{H\times W\times 2C}$, we firstly shrink it using a $1\times1$ convolutional layer to get the feature $\tilde{\textrm{F}}\in\mathbb{R}^{H\times W\times C}$. Then $\tilde{\textrm{F}}$ is reshaped to $\mathbf{X}\in \mathbb{R}^{HW\times C}$. We utilize matrix multiplication to compute pair-wise similarities of inner product, and obtain the affinity matrix $\mathbf{M} = \mathbf{X}\mathbf{X}^\intercal\in \mathbb{R}^{HW\times HW}$.
Other pair-wise functions such as $e^{-\|\textrm{X}_{i} - \textrm{X}_{j}\|}$ can also be used, just not shown in the figure.
Note that, different from non-local blocks \cite{wang2017non}, our affinity matrix must satisfy the symmetric and nonnegative properties to represent the pair-wise similarities.
Finally, as each row of the matrix $\mathbf{M}$ represents the pair-wise relationships between one position and all other positions, we conduct normalization along each row of $\mathbf{M}$ to reduce the influence of scale. In this way, the task-level patterns can be represented in each $\mathbf{M}$. Note that we add no supervision to learn $\mathbf{M}$ as literature \cite{bertasius2017convolutional}, because such supervision will cost extra memories and be not easy to define for some tasks.
After that, we want to integrate the cross-task information for each task. Denote these three tasks as $T_1, T_2, T_3$, and the corresponding affinity matrices as $ \mathbf{M}_{T_1}$£¬$\mathbf{M}_{T_2}$£¬$\mathbf{M}_{T_3}$, then we can learn weights $\alpha^{T_i}_k$ ($k = 1,2,3, \sum_{k=1}^{n} \alpha^{T_i}_k=1$) to adaptively combine the matrices as:
\begin{equation}\label{eqn6}
\hat{\mathbf{M}}_{T_i} = \alpha^{T_i}_1 \cdot \mathbf{M}_{T_1} + \alpha^{T_i}_2 \cdot \mathbf{M}_{T_2} + \alpha^{T_i}_3 \cdot \mathbf{M}_{T_3}.
\end{equation}
In this way, the cross-task affinitive patterns can be propagated into $\hat{\mathbf{M}}_{T_i}$.
In practice, we implement affinity learning layers at decoding process on 1/16, 1/8 and 1/4 input scale respectively, hence it actually learns non-local patch-level relationships.

\subsection{Task-Specific Propagation}

After obtaining the combined affinity matrices, we spread such affinitive patterns into the feature space of each task by the task-spacific propagation. Different from non-local block \cite{wang2017non} and local spatial propagation \cite{liu2017learning, cheng2018depth},
we perform an iterative non-local diffusion process in each diffusion layer to capture long-distance similarities, as illustrated in Fig.~\ref{fig3}(b). The diffusion process is performed on initial prediction as well as features from task-specific network. Without loss of generality, assuming feature or initial prediction $\textrm{P} \in \mathbb{R}^{H\times W\times C}$ is from task-specific network, we firstly reshape it to $\mathbf{h}\in \mathbb{R}^{HW\times C}$, and perform one step diffusion by using matrix multiplication with $\hat{\mathbf{M}}$. In this way, the feature vector of each position is obtained by weighted accumulating feature vectors of all positions using the learned affinity. Note that such one-step diffusion may not deeply and effectively propagate the affinity information to the feature space, we perform the multi-step iterative diffusion as:
\begin{equation}\label{iterdiff}
\mathbf{h}^{t+1} = \hat{\mathbf{M}} \mathbf{h}^t,\ \ t\geq0,
\end{equation}
where $\mathbf{h}^t$ means the diffused feature (or prediction) at step $t$. Such diffusion process can be also expressed with a partial differential equation (PDE):
\begin{equation}\label{pde}
    \begin{aligned}
        \mathbf{h}^{t+1} =& \ \ \hat{\mathbf{M}} \mathbf{h}^t =(\mathbf{I} - \mathbf{L})\mathbf{h}^t,\\
        \mathbf{h}^{t+1} - \mathbf{h}^t =& -\mathbf{L} \mathbf{h}^t,\\
        \partial_t \mathbf{h}^{t+1} =& -\mathbf{L} \mathbf{h}^t,
    \end{aligned}
\end{equation}
where $\mathbf{L}$ is the Laplacian matrix. As $\hat{\mathbf{M}}$ is normalized and has finite Frobenius norm, the stability of such PDE can be guaranteed \cite{liu2017learning}. Assuming we totally perform $t^*$ steps in each diffusion layer, in order to prevent the feature deviating too much from the initial one, we use the weighted accumulation on the initial feature (or prediction) $\mathbf{h}^0$ as:
\begin{equation}\label{finaldiff}
\mathbf{h}^{\textrm{out}} = \beta \mathbf{h}^{t^*} + (1-\beta)\mathbf{h}^0,\ \ 0\leq \beta \leq1,
\end{equation}
where $\mathbf{h}^{\textrm{out}}$ means the final output from a diffusion layer. In this way, the learned affinitive patterns in each $\hat{\mathbf{M}}_{T_i}$ can be effectively propagated into each task $T_i$.

\subsection{The Loss Function}
In this section we introduce a pair-wise affinity loss for our PAP network. As PAP method is designed to learn task-correlative pair-wise similarities, we also hope our loss function can enhance the pair-wise constraints. Firstly we define the prediction at position $i$ is $\hat{z}_i$, and the corresponding ground truth is $z_i$. Then we define the pair-wise distance in prediction and corresponding ground truth as $\hat{d}_{ij} = |\hat{z}_i-\hat{z}_j|$ and $d_{ij} = |z_i-z_j|$. We hope the distance in prediction to be similar to ground truth, so the pair-wise loss can be defined as $\mathcal{L}_{\textrm{pair-wise}} = \sum_{\forall i,j} |\hat{d}_{ij} - d_{ij}|$. As the calculation of the pair-wise loss in each task will have a high memory burden, so we randomly select $S$ pairs from each task and then compute the pair-wise loss $\mathcal{L}_{\textrm{pair-wise}} = \sum_{S} |\hat{d}_{ij} - d_{ij}|$. As the pairs are randomly selected, such pair-wise loss can capture similarities of various-distance pairs, not only the adjacent pixels in \cite{eigen2015predicting}. Meanwhile, we also use berHu loss \cite{Laina2016Deeper}, L1 loss and cross-entropy loss for depth estimation, surface normal prediction and semantic segmentation respectively, which are denoted as $\mathcal{L}^{T_i}$ ($T_i$ means the $i$-th task). Finally the total loss of the joint task learning problem can be defined as:
\begin{equation}\label{loss}
\mathcal{L} = \sum_{T_i} \lambda_{T_i} (\mathcal{L}^{T_i} + \xi_{T_i}\mathcal{L}^{T_i}_{\textrm{pair-wise}}),
\end{equation}
where $\mathcal{L}^{T_i}_{\textrm{pair-wise}}$ is the pair-wise loss for the corresponding $i$-th task, and $\lambda_{T_i}$ and $\xi_{T_i}$ are two weights for the $i$-th task.

\section{Experiment}
\subsection{Dataset}
\textbf{NYUD-v2:} The NYUD v2 dataset~\cite{silberman2012indoor} consists of RGB-D images of 464 indoor scenes. There are 1449 images with semantic labels, 795 of them are used for training and the remaining 654 images for testing. We randomly select more images (12k, same as \cite{Laina2016Deeper, zhang2018joint} ) from the raw data of official training scenes. These images have the corresponding depth maps but no semantic labels or surface normals. We follow the procedure in \cite{fouhey2013data} and \cite{qi2018geonet} to generate surface normal ground truth. In this way, we can use more data to train our model for jointly depth and surface normal prediction.

\textbf{SUN RGBD:} The SUN RGBD dataset~\cite{Song2015SUN} contains 10355 RGBD images with semantic labels of which 5285 for training and 5050 for testing. We use the official training set with depth and semantic labels to train our network, and the official testing set for evaluation. There is no surface normal ground truth on this dataset, so we perform experiments on jointly predicting depth and segmentation on this dataset. 

\textbf{KITTI:} KITTI online benchmark~\cite{uhrig2017sparsity} is a widely-used outdoor dataset for depth estimation. There are 4k images for training, 1k images for validating and 500 images for testing on the online benchmark. As it has no semantic labels or surface normal ground truth, we mainly transform such information using our PAP method to demonstrate that PAP can distilling knowledge to improve the performance.

\subsection{Implementation Details and Metrics}
We implement the proposed model using Pytorch \cite{paszke2017automatic} on a single Nvidia P40 GPU. We build our network based on ResNet-18 and ResNet-50, and each model is pre-trained on the ImageNet classification task~\cite{deng2009imagenet}. In diffusion process, we use a same subsampling strategy as \cite{wang2017non} to downsample $h$ in Eqn.~(\ref{iterdiff}), which can reduce the amount of pair-wise computation by 1/4. We set the trade-off parameter $\beta$ to 0.05. 300 pairs are randomly selected to compute the pair-wise loss in each task. We simply set $\lambda_{T_i} = \frac{1}{3}$ and $\xi_{T_i} = 0.2$ to balance the loss functions. Initial learning rate is set to $10^{-4}$ for the pre-trained convolutional layers and 0.01 for the other layers. For NYUD-v2, we train the model of 795 training images for 200 epochs and fine-tune 100 epochs, and train the model of 12k training images for jointly depth/normal predicting for 30 epochs and fine-tune for 10 epochs.
For SUN-RGBD dataset, we train the model for 30 epochs and fine-tune it for 30 epochs using a learning rate of 0.001. For KITTI, we first train the model on NYUD-v2 for surface normal estimation, and then freeze the surface normal branch to train depth branch on KITTI for 15 epochs, finally we freeze the normal branch and fine-tune the model on KITTI for 20 epochs.

Similar to the previous works~\cite{Laina2016Deeper,eigen2015predicting,Xu2017CVPR}, we evaluate our depth prediction results with the root mean square error (rmse), average relative error (rel), root mean square error in log space (rmse-log), and accuracy with threshold ($\delta$): \% of $\widetilde{x_i}$ s.t. max($\frac{\widetilde{x_i}}{x_i}$, $\frac{x_i}{\widetilde{x_i}}$)=$\delta$, $\delta = 1.25, 1.25^2, 1.25^3$,
where $\widetilde{x_i}$ is the predicted depth value at the pixel $i$, $n$ is the number of valid pixels and $x_i$ is the ground truth. The evaluation metrics for surface normal prediction \cite{wang2015designing, bansal2016marr, eigen2015predicting} are mean of angle error (mean), medians of the angle error (median), root mean square error for normal (rmse-n \%), and pixel accuracy as percentage of pixels with angle error below threshold $\eta$ where $\eta \in [11.25^\circ, 22.50^\circ, 30^\circ]$. For the evaluation of semantic segmentation results, we follow the recent works ~\cite{Cheng2017Locality} \cite{kendall2015bayesian} \cite{Lin2016RefineNet} and use the common metrics including pixel accuracy (pixel-acc), mean accuracy (mean-acc) and mean intersection over union (IoU).

\subsection{Ablation Study}
In this section we perform many experiments to analyse the influence of different settings in our method.

\begin{table}[!t]
\caption{Analyses on Joint task learning on NYU Depth V2.}
  \begin{center}
      \footnotesize
    \begin{tabular}[0.618\textwidth]{lccc}
    \hline

  Metric&rmse&iou&rmse-n\\
  \hline
  Depth only &0.570&-&\\
  Segmentation only& -&42.8& -\\
  Normal only&-&-&28.7\\
  Depth\&Seg jointly&0.556&44.3&-\\
  Depth\&Normal jointly&0.550&-&28.1\\
  Segmentation\&Normal jointly&-&44.5&28.3\\
  Three task jointly&\textbf{0.533}&\textbf{46.2}&\textbf{26.9}\\

\hline
    \end{tabular} 
    \end{center}
  \label{table1} \vspace{-0.2cm}
\end{table}%
\textbf{Effectiveness of joint task learning:} We first analyse the benefit of joint predicting depth, surface normal and semantic segmentation using our PAP method. The networks are trained on NYUD v2 dataset, and we select ResNet-18 as our shared network backbone and only learn the affinity matrix on 1/8 input scale in each experiment.
As illustrated in Table~\ref{table1}, we can see that joint-task models gets superior performances than the single task model, and further jointly learning three tasks obtains best results. It can be revealed that our PAP method does boost each task in the jointly learning procedures.

\begin{table}[!t]
\caption{Comparisons of different network settings and baselines on NYU Depth v2 dataset. }
  \begin{center}
      \footnotesize
    \begin{tabular}[0.618\textwidth]{lccc}
    \hline Method
    &rmse&IoU&rmse-n\\
\hline
  initial prediction
  &0.582&41.3&29.6\\
  + PAP w/o cross-t prop.
  &0.574&41.8&29.1\\
  + PAP cross-t prop.
  &0.558&43.1&28.5\\
  + PAP cross-t prop. + recon-net
  &0.550&43.8&28.2\\
  + PAP cross-t prop + recon-net + pair-loss
  &0.543&44.2&27.8\\
  + cross-stich~\cite{Misra2016Cross}&0.550&43.5&28.2\\
  + CSPN~\cite{cheng2018depth}&0.548&43.8&28.0\\
  \hline
  aff-matrix on 1/16 input scale
  &0.543&44.2&27.8\\
  aff-matrix on 1/8 input scale
  &0.533&\textbf{46.2}&26.9\\
  aff-matrix on 1/4 input scale
  &\textbf{0.530}&46.5&\textbf{26.7}\\
\hline
  Inner product& 0.543&44.2&27.8\\
  L1 distance&0.540&44.0&27.9\\
  \hline
    \end{tabular} 
    \end{center}
  \label{t2} \vspace{-0.7cm}
\end{table}
\textbf{Analysis on network settings:} We perform many experiments to analyse the effectiveness of each network modules. In each experiment we use ResNet-18 as our network backbone for equally comparing, and each model is trained on NYUD v2 dataset for the three tasks. The result can be seen in Table~\ref{t2}. Note that the results of first five rows are computed from the model with affinity matrix learned on 1/16 input scale. We can observe that PAP, reconstruction net and pair-wise loss can all contribute to improve the performance. We also compare two approaches in the same settings, i.e., cross-stich units~\cite{Misra2016Cross} and convolutional spatial propagation layers~\cite{cheng2018depth} which can also fuse and interact cross-task information. We find that they obtain weaker performances. It may be attributed to that: a) cross-stich layer only combines features, but cannot represent the affinitive patterns between tasks; b) they only use limited local information.
The middle three rows of the Table~\ref{t2} show the influence on which scale the affinity matrix is learned. We can find that learning affinity matrix on a larger scale may be beneficial, as the larger affinity matrices can describe the similarities between more patches. Note that the improvements of learning matrix on 1/4 input scale are comparatively smaller, and the reason may be that learning good non-local pair-wise similarities becomes more difficult with scale increasing. Finally we show the results using different functions to calculate the similarities. We find that these two functions does produce different performances, but with little difference. Hence, we mainly use dot product as our weight function in the following experiments for convenience.

\begin{figure}[!t]
  \centering
\includegraphics[width=7.5cm]{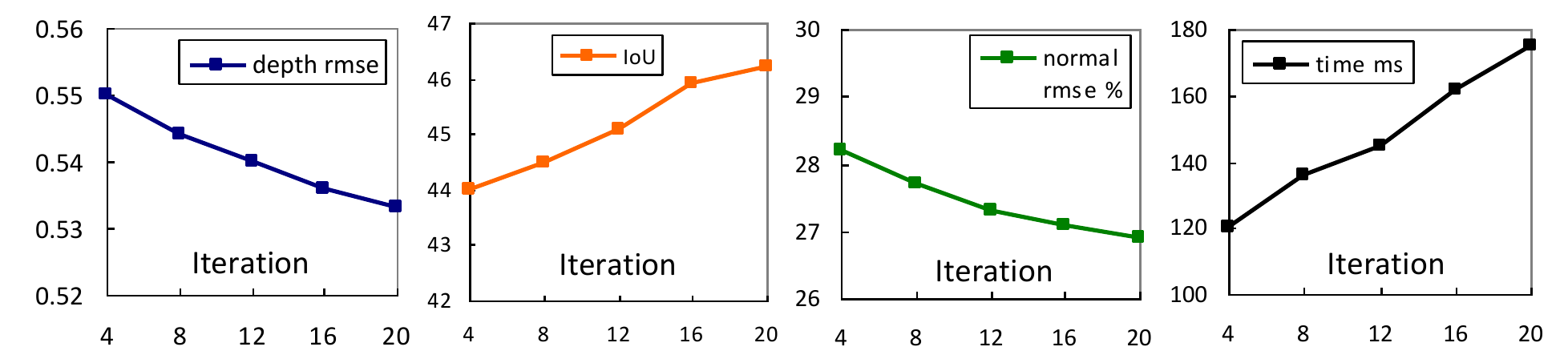}
   \caption{The influence of the iterations in diffusion process. The performance and time burden changes can be seen as a trade-off.
   }\label{fig4}
  \vspace{-0.3cm}
\end{figure}
\textbf{Influence of the iteration:} Here we make experiments to analyse the influence of the iterative steps in Eqn.~(\ref{iterdiff}). The models are based on ResNet-18 and trained on NYUD v2 dataset, and the affinity matrices are learned on 1/8 input scale. While testing, the input size is 480$\times$640. As illustrated in Fig.~\ref{fig4}, we can see that the performances of all tasks are improved with more iterations, at least in such a range. These results demonstrate that the pair-wise constraints and regularization may be enhanced with more iterations in diffusion. But the testing time will also increase with more steps, which can be seen as a trade-off.

\begin{figure}[!t]
  \centering
\includegraphics[width=7.5cm]{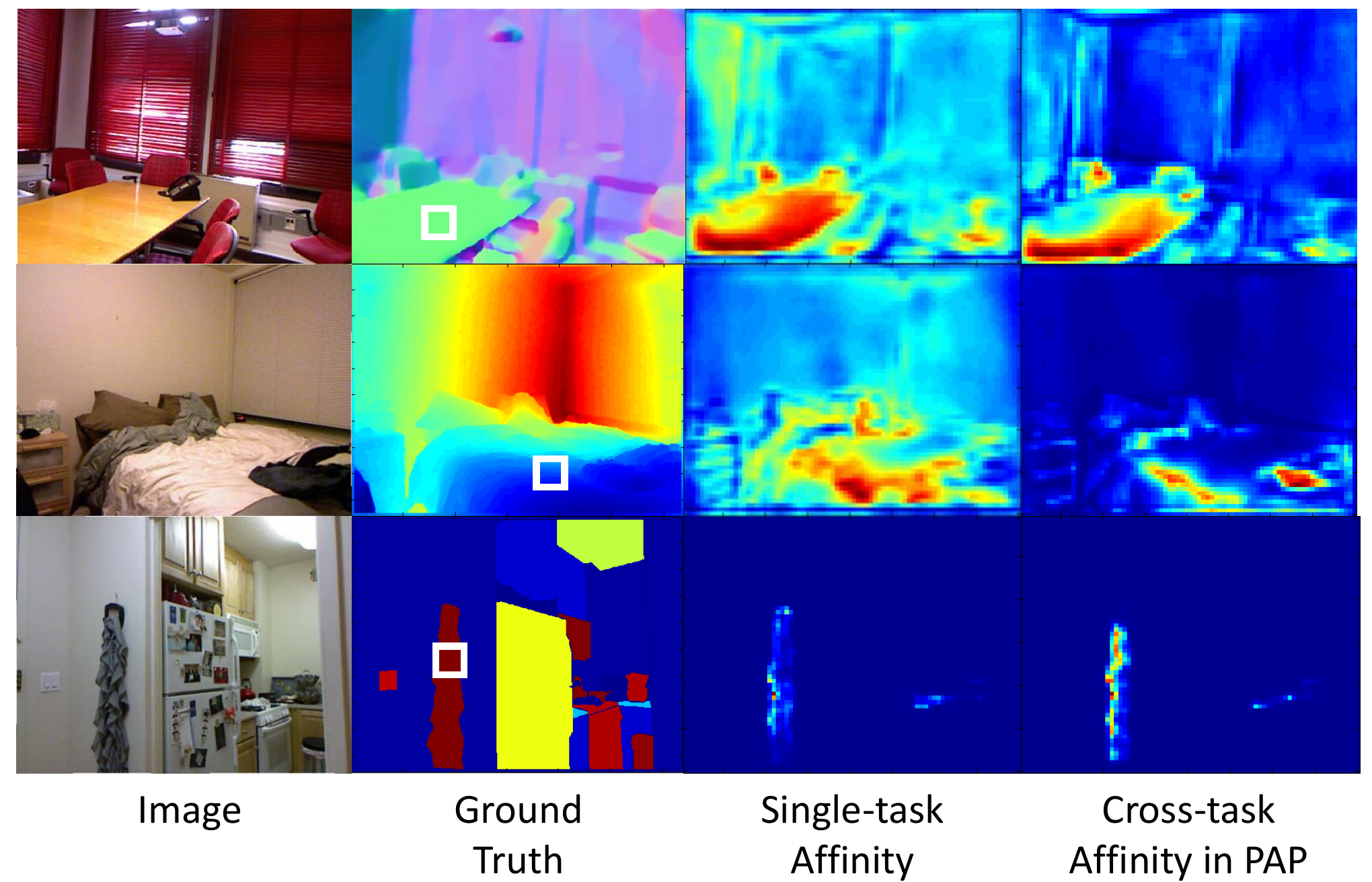}
   \caption{Visualization of the single-task and our cross-task affinity maps at the white point for each task. We can see that the pair-wise similarities at the white point can be improved and corrected in our PAP method.
   }\label{fig5}
  \vspace{-0.2cm}
\end{figure}
\textbf{Visualization of the affinity matrices:} We show several examples of the learned affinity maps in Fig.~\ref{fig5}. Note that the affinity maps belong to the white point in each image. We can see that the single-task affinity maps often show improper pair-wise relationships, while the cross-task affinity maps in our PAP method have closer relationships with the points which have similar depth, normal direction and semantic label. As the affinity matrices is non-local and actually a dense graph, it can well represent the long-distance similarities. Such observations demonstrate that the cross-task complementary affinity information can be learned to refine the single-task similarities in PAP method.
Though without supervision as \cite{bertasius2017convolutional}, our PAP method can still learn good affinity matrices in such task-regularized unsupervised approach.

\subsection{Comparisons with state-of-the-art methods}
\begin{table}[!t]
\caption{Comparisons with the state-of-the-art depth estimation approaches on NYU Depth V2 Dataset.} \vspace{-0.25cm}
  \begin{center}
      \scriptsize
    \begin{tabular}[0.618\textwidth]{lc|cccccc}
    \hline Method&
    data&rmse&rel&log&$\delta_1$&$\delta_2$&$\delta_3$\\
\hline

  HCRF~\cite{li2015depth}&795 & 0.821 & 0.232 & - &0.621 &0.886 &0.968 \\

 DCNF~\cite{liu2015Learning}&795 & 0.824 & 0.230 & - &0.614 &0.883 &0.971\\

 Wang~\cite{Wang2015Towards}&795 & 0.745 & 0.220 & 0.262 & 0.605 &0.890 & 0.970\\
 NR forest~\cite{Roy2016CVPR}&795 & 0.744 & 0.187 & - &-&-&-\\
 Xu~\cite{xu2018structured}&795&0.593&0.125&-&0.806&0.952&0.986\\
 PAD-Net~\cite{xu2018pad} &795 &0.582&0.120&-&0.817&0.954&0.987\\
\hline
 Eigen~\cite{eigen2014depth}&120k & 0.877 & 0.214 & 0.285  &0.611 & 0.887 & 0.971 \\

 MS-CNN~\cite{eigen2015predicting}&120k & 0.641 & 0.158 & 0.214  & 0.769 &0.950 &0.988 \\
 MS-CRF~\cite{Xu2017CVPR}&95k & 0.586 & 0.121 & - & 0.811 & 0.954 & 0.987 \\

 FCRN~\cite{Laina2016Deeper}&12k & 0.573 & 0.127 & 0.194 &0.811 &0.953 &0.988 \\

 GeoNet~\cite{qi2018geonet}&16k &0.569&0.128&-&0.834&0.960&0.990\\
 AdaD-S~\cite{Kundu_2018_CVPR}&100k&0.506&\textbf{0.114}&-&\textbf{0.856}&0.966&0.991\\

 DORN~\cite{fu2018deep}&120k&0.509&0.115&-&0.828&0.965&0.992\\
 TRL~\cite{zhang2018joint}&12k&0.501&0.144&0.181&0.815&0.962&0.992\\
\hline
 Ours d+s+n &795 &0.530&0.142&0.190&0.818&0.957&0.988\\
 Ours d+n &12k & \textbf{0.497} & 0.121 & \textbf{0.175} &0.846&\textbf{0.968}&\textbf{0.994} \\

\hline
    \end{tabular}
    \end{center}
  \label{t3} \vspace{-0.4cm}
\end{table}%
\begin{figure}[!t]
  \centering
\includegraphics[width=8cm]{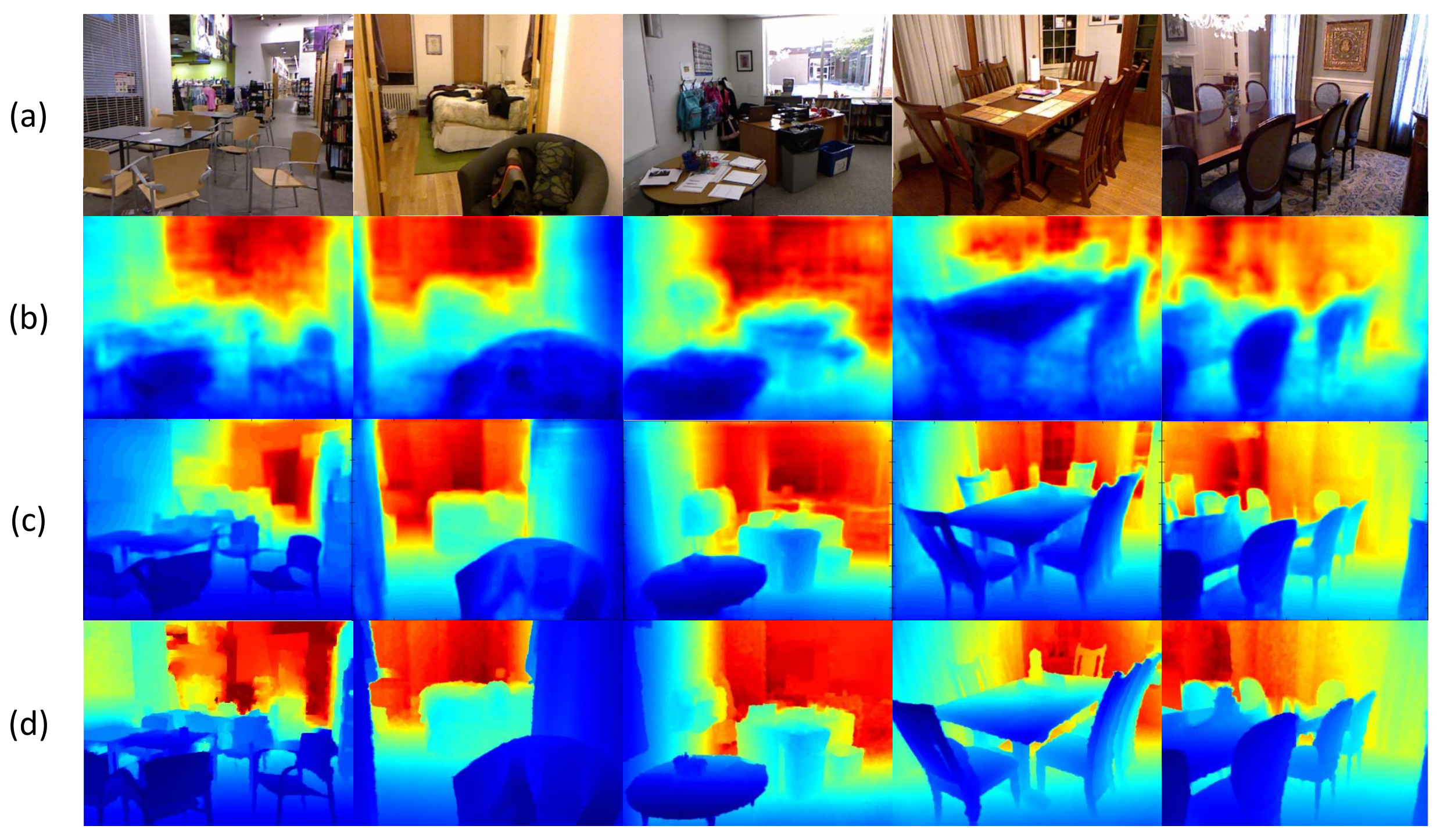}
   \caption{Visualization of our predicted depth maps. (a) image; (b) predictions of \cite{xu2018structured}; (c) our results; (d) ground truth. We can find that our predictions have obviously finer details and closer to ground truth.
   }\label{fig6}
  \vspace{-0.4cm}
\end{figure}
\textbf{Depth Estimation:} We mainly perform experiments on NYUD-v2 dataset to evaluate our depth predictions. The models are based on ResNet-50.
As illustrated in Table~\ref{t3}, our model trained for three tasks (ours d+s+n) obtains competitive results, though only 795 images are used for training. Such results demonstrate that our PAP method can well boost each task and benefit joint task learning with limited training data. For the model trained for depth\&normal prediction (ours d+n), with more training data can be used, our PAP method gets significantly best performances in most of the metrics with more training data, which well proves the effectiveness of our approach. Qualitative results can be observed in Fig.~\ref{fig6}, compared with the recent work \cite{xu2018structured}, our predictions are more fine-detailed and closer to ground truth.

\begin{table}[!t]
\caption{Comparisons with the state-of-the-art surface normal estimation approaches on NYU Depth V2 Dataset.} 
  \begin{center}
      \scriptsize
    \begin{tabular}[0.618\textwidth]{lcccccc}
    \hline Method
    &mean&median&rmse-n&$11.25^\circ$&$22.50^\circ$&$30^\circ$\\
\hline

  3DP~\cite{fouhey2013data} & 36.3 & 19.2 & - &16.4 &36.6 &48.2 \\

 UNFOLD~\cite{fouhey2014unfolding} & 35.2 & 17.9 & - &40.5 &54.1 &58.9\\

 Discr.~\cite{zeisl2014discriminatively} & 33.5 & 23.1 & - & 27.7 &49.0 & 58.7\\

 MS-CNN~\cite{eigen2015predicting} & 23.7 & 15.5 & -  &39.2 & 62.0 & 71.1 \\

 Deep3D~\cite{wang2015designing} & 26.9 & 14.8 & - &42.0&61.2&68.2\\

 SkipNet~\cite{bansal2016marr} & 19.8 & 12.0 & 28.2  & 47.9 &70.0 &77.8 \\
 SURGE~\cite{wang2016surge} & 20.6 & 12.2 & - & 47.3 & 68.9 & 76.6 \\
 GeoNet~\cite{qi2018geonet} &19.0&11.8&26.9&48.4&71.5&79.5\\
\hline
 Ours-VGG16 & \textbf{18.6} & \textbf{11.7} & \textbf{25.5} &\textbf{48.8}&\textbf{72.2}&\textbf{79.8} \\

\hline
    \end{tabular}
    \end{center}
  \label{t4} \vspace{-0.3cm}
\end{table}%
\begin{figure}[!t]
  \centering
\includegraphics[width=7.5cm]{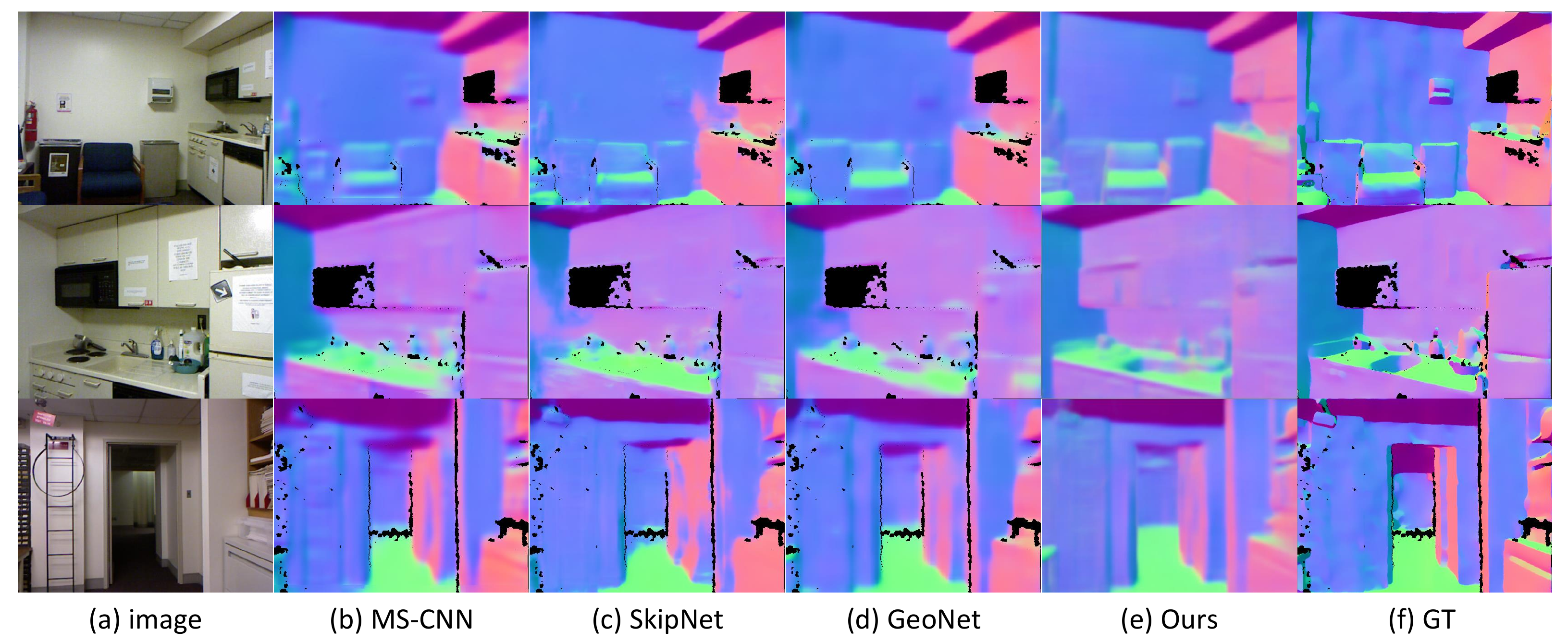}
   \caption{Visualization of our predicted surface normal. (a) image; (b) predictions of \cite{eigen2015predicting}; (c) predictions of~\cite{bansal2016marr} ; (d) predictions of \cite{qi2018geonet}; (e) our results; (f) ground truth.
   }\label{fig7}
  \vspace{-0.3cm}
\end{figure}
\textbf{Surface Normal Estimation:} We mainly evaluate our surface normal predictions on NYUD-v2 dataset. As previous methods mainly build their network based on VGG-16~\cite{simonyan2014very}, we also utilize the same setting in our experiments. As illustrated in Table~\ref{t4}, our PAP method obtains obviously superior performances than the previous approaches in all metrics. Such results well demonstrate that our joint task learning method can boost and benefit the surface normal estimation. Qualitative results can be observed in Fig.~\ref{fig7}, we can find that our method can produce better or competitive results.

\begin{table}[!t]
\caption{Comparisons the state-of-the-art semantic segmentation methods on NYU Depth v2 dataset.}
  \begin{center}
      \scriptsize
    \begin{tabular}[0.618\textwidth]{lcccc}
    \hline Method
    &data&pixel-acc&mean-acc&IoU\\
\hline
  FCN~\cite{Long2017Fully} &RGB&60.0&49.2&29.2\\
  Context~\cite{Lin2016Efficient}
  &RGB&70.0&53.6&40.6\\
  Eigen \textit{et al.}~\cite{eigen2015predicting}
  &RGB&65.6&45.1&34.1\\
  B-SegNet~\cite{kendall2015bayesian}
  &RGB&68.0&45.8&32.4\\
  RefineNet-101~\cite{Lin2016RefineNet}
  &RGB&72.8&57.8&44.9\\
  PAD-Net~\cite{xu2018pad}&RGB&75.2&62.3&50.2\\
  TRL-ResNet50~\cite{zhang2018joint}&RGB&\textbf{76.2}&56.3&46.4\\
  \hline
  Deng \textit{et al.}~\cite{Deng2015Semantic}
  &RGBD&63.8&-&31.5\\
  He \textit{et al.}~\cite{He2016STD2P}
  &RGBD&70.1&53.8&40.1\\
  LSTM~\cite{Li2016LSTM}
  &RGBD&-&49.4&-\\
  Cheng \textit{et al.}~\cite{Cheng2017Locality}
  &RGBD&71.9&60.7&45.9\\
  3D-GNN~\cite{qi20173d}
  &RGBD&-&55.7&43.1\\
  RDF-50~\cite{SJ2017RDF}
  &RGBD&74.8&60.4&47.7\\
  \hline
  Ours-ResNet50 &RGB&\textbf{76.2}&\textbf{62.5}&\textbf{50.4}\\

\hline
    \end{tabular}
    \end{center}
  \label{t5} \vspace{-0.7cm}
\end{table}
\begin{table}[!t]
\caption{Comparison with the state-of-the-art semantic segmentation methods on SUN-RGBD dataset. }
  \begin{center}
      \scriptsize
    \begin{tabular}[0.618\textwidth]{lcccc}
    \hline Method
    &data&pixel-acc&mean-acc&IoU\\
\hline
  Context~\cite{Lin2016Efficient}
  &RGB&78.4&53.4&42.3\\
  B-SegNet~\cite{kendall2015bayesian}
  &RGB&71.2&45.9&30.7\\
  RefineNet-101~\cite{Lin2016RefineNet}
  &RGB&80.4&57.8&45.7\\
  TRL-ResNet50~\cite{zhang2018joint}
  &RGB&83.6&58.9&50.3\\
  \hline
  LSTM~\cite{Li2016LSTM}
  &RGBD&-&48.1&-\\
  Cheng \textit{et al.}~\cite{Cheng2017Locality}
  &RGBD&-&58.0&-\\
  CFN~\cite{Di2017CFN}
  &RGBD&-&-&48.1\\
  3D-GNN~\cite{qi20173d}
  &RGBD&-&57.0&45.9\\
  RDF-152~\cite{SJ2017RDF}
  &RGBD&81.5&\textbf{60.1}&47.7\\
  \hline

  Ours-ResNet50 &RGB&\textbf{83.8}&58.4&\textbf{50.5}\\

\hline
    \end{tabular}
    \end{center} \vspace{-0.6cm}
  \label{t6}
\end{table}
\begin{figure}[!t]
  \centering
\includegraphics[width=7.5cm]{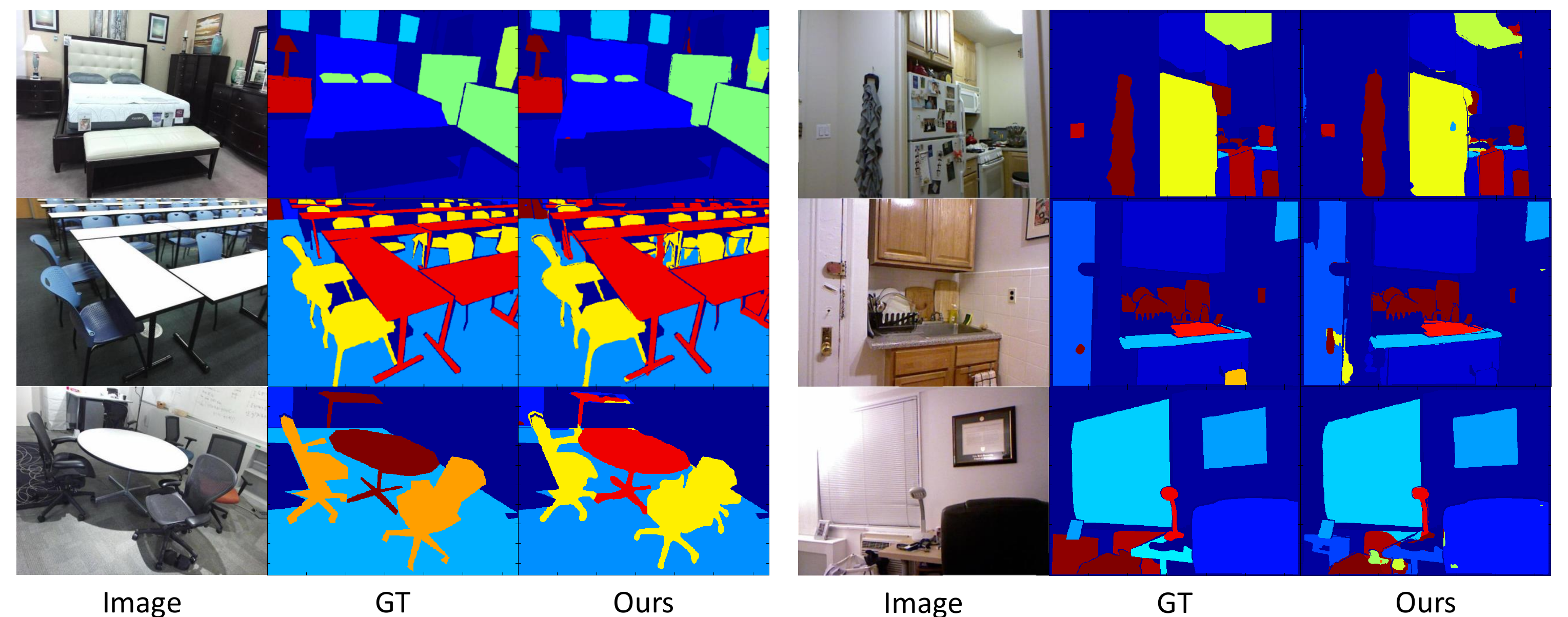}
   \caption{Qualitative semantic segmentation results of our method on NYUD-v2 and SUNRGBD datasets.
   }\label{fig8}
  \vspace{-0.3cm}
\end{figure}
\textbf{RGBD Semantic Segmentation:} We evaluate our segmentation results on widely-used NYUD-v2 and SUN-RGBD datasets. The model in each experiment is build based on ResNet-50 and trained for the three tasks on NYUD-v2, and jointly depth prediction and semantic segmentation on SUN-RGBD. The performance on NYUD-v2 dataset is shown in Table~\ref{t5}. We can observe that the performances of our PAP method are superior or competitive, though using only RGB images as input. Such results can demonstrate that although depth ground truth is not directly use, our method can benefit the segmentation from jointly learning depth information. The performances on SUN-RGBD dataset are illustrated in Table~\ref{t6}, we can see that though slightly weaker than RDF-152~\cite{SJ2017RDF} in mean-acc metric, our method can obtain best results in other metrics. Such results reveal that our predictions are superior or at least competitive with state-of-the-art methods. Visualized results can be observed in Fig.~\ref{fig8}, we can see that our predictions are with high quality and close to ground truth.

\begin{table}[!t]
\caption{Comparison with the state-of-the-art methods on KITTI online benchmark (lower is better). }
  \begin{center}
      \scriptsize
    \begin{tabular}[0.618\textwidth]{lccccc}
    \hline Method
    &SILog&sqErrRel&absErrRel&iRMSE&time\\
\hline
  DORN~\cite{fu2018deep}&11.77&2.23&8.78&12.98&0.5s\\
  VGG16-Unet$^*$&13.41&2.86&10.60&15.06&0.16s\\
  FUSION-ROB$^*$&13.90&3.14&11.04&15.69&2s\\
  BMMNet$^*$&14.37&5.10&10.92&15.51&0.1s\\
  DABC~\cite{li2018deep}&14.49&4.08&12.72&15.53&0.7s\\
  APMoE~\cite{kong2018pixel}&14.74&3.88&11.74&15.63&0.2s\\
  CSWS~\cite{li2018monocular}&14.85&3.48&11.84&16.38&0.2s\\
  \hline
  Ours single&14.58&3.96&11.50&15.24&0.1s\\
  Ours cross-stich~\cite{Misra2016Cross}&14.33&3.85&11.23&15.14&0.1s\\
  Ours&13.08&2.72&10.27&13.95&0.2s\\
\hline
    \end{tabular}
    \end{center} \vspace{-0.8cm}
  \label{t7}
\end{table}
\begin{figure}[!t]
  \centering
\includegraphics[width=7.5cm]{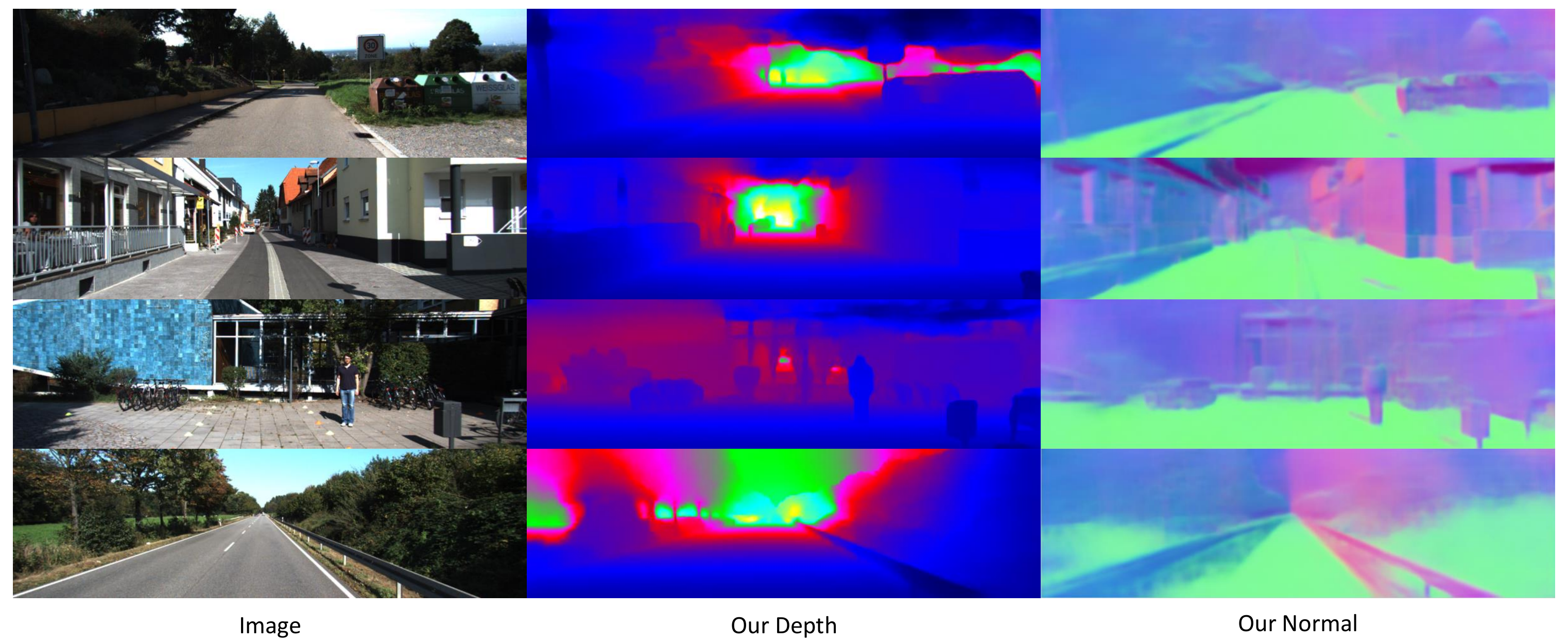}
   \caption{Qualitative results of our method on KITTI dataset. We can find that our model obtains good depth predictions and normal estimations.
   }\label{fig9}
  \vspace{-0.5cm}
\end{figure}
\subsection{Effectiveness On Distilling}
Sometimes the ground truth data cannot be always available for each task, e.g., some widely-used outdoor depth datasets, such as KITTI~\cite{uhrig2017sparsity}, has no or very limited surface normal and segmentation ground truth. However, we can use PAP method to distill the knowledge from other dataset to boost the target task. We train our model on NYUD-v2 for depth and normal estimation, and then freeze the normal branch to train the model on KITTI. We evaluate our predictions on the KITTI online evaluation server, and the results are shown in Table~\ref{t7} ($^*$ means anonymous method). Our PAP method outperforms our single-task and cross-stich based model. Compared with the state-of-the-art methods, though slightly weaker than DORN~\cite{fu2018deep}, our method obtains superior performances than all other published or unpublished approaches. Note that our method runs faster than DORN, which can be seen as a trade-off. These results demonstrate the effectiveness and potential of PAP method on task distilling and transferring. Qualitative results can be seen on Fig.~\ref{fig9}, and our predictions on depth and normal are both with high quality.

\section{Conclusion}
In this paper, we propose a novel Pattern-affinitive Propagation method for jointly predicting depth, surface normal and semantic segmentation. Statistic results have shown that the affinitive patterns among tasks can be modeled in pair-wise similarities to some extent. The PAP can effectively learn the pair-wise relationships from each task, and further utilize such cross-task complementary affinity to boost and regularize the joint task learning procedure via the cross-task and task-specific propagation. Extensive experiments demonstrate our PAP method obtained state-of-the-art or competitive results on these three tasks.
In the future, we may generalize and improve the efficiency of the method on more vision tasks.

\section{Acknowledgement}
This work was supported by the National Science Fund of China under Grant Nos. U1713208, 61806094, 61772276 and 61602244, Program for Changjiang Scholars, and ¡°111¡± Program AH92005.
{\small
\bibliographystyle{ieee_fullname}
\bibliography{egbib}

\begin{thebibliography}{10}\itemsep=-1pt

\bibitem{bansal2016marr}
Aayush Bansal, Bryan Russell, and Abhinav Gupta.
\newblock Marr revisited: 2d-3d alignment via surface normal prediction.
\newblock In {\em CVPR}, pages 5965--5974, 2016.

\bibitem{bertasius2017convolutional}
Gedas Bertasius, Lorenzo Torresani, X~Yu Stella, and Jianbo Shi.
\newblock Convolutional random walk networks for semantic image segmentation.
\newblock In {\em CVPR}, pages 6137--6145, 2017.

\bibitem{Caruana1997Multitask}
Rich Caruana.
\newblock Multitask learning.
\newblock {\em Machine Learning}, 28(1):41--75, 1997.

\bibitem{chen2015deepdriving}
Chenyi Chen, Ari Seff, Alain Kornhauser, and Jianxiong Xiao.
\newblock Deepdriving: Learning affordance for direct perception in autonomous
  driving.
\newblock In {\em ICCV}, pages 2722--2730, 2015.

\bibitem{cheng2018depth}
Xinjing Cheng, Peng Wang, and Ruigang Yang.
\newblock Depth estimation via affinity learned with convolutional spatial
  propagation network.
\newblock In {\em ECCV}, pages 108--125, 2018.

\bibitem{Cheng2017Locality}
Yanhua Cheng, Rui Cai, Zhiwei Li, Xin Zhao, and Kaiqi Huang.
\newblock Locality-sensitive deconvolution networks with gated fusion for rgb-d
  indoor semantic segmentation.
\newblock In {\em CVPR}, volume~3, pages 1475--1483, 2017.

\bibitem{deng2009imagenet}
Jia Deng, Wei Dong, Richard Socher, Li-Jia Li, Kai Li, and Li Fei-Fei.
\newblock Imagenet: A large-scale hierarchical image database.
\newblock In {\em CVPR}, pages 248--255, 2009.

\bibitem{Deng2015Semantic}
Zhuo Deng, Sinisa Todorovic, and Longin~Jan Latecki.
\newblock Semantic segmentation of rgbd images with mutex constraints.
\newblock In {\em ICCV}, pages 1733--1741, 2015.

\bibitem{Di2017CFN}
L Di, Chen Guangyong, Cohen-Or Daniel, Heng Pheng-Ann, and Huang Hui.
\newblock Cascaded feature network for semantic segmentation of rgb-d images.
\newblock In {\em ICCV}, pages 1320--1328, 2017.

\bibitem{eigen2015predicting}
David Eigen and Rob Fergus.
\newblock Predicting depth, surface normals and semantic labels with a common
  multi-scale convolutional architecture.
\newblock In {\em ICCV}, pages 2650--2658, 2015.

\bibitem{eigen2014depth}
David Eigen, Christian Puhrsch, and Rob Fergus.
\newblock Depth map prediction from a single image using a multi-scale deep
  network.
\newblock In {\em NIPS}, pages 2366--2374, 2014.

\bibitem{fong2003survey}
Terrence Fong, Illah Nourbakhsh, and Kerstin Dautenhahn.
\newblock A survey of socially interactive robots.
\newblock {\em Robotics and autonomous systems}, 42(3-4):143--166, 2003.

\bibitem{fouhey2013data}
David~F Fouhey, Abhinav Gupta, and Martial Hebert.
\newblock Data-driven 3d primitives for single image understanding.
\newblock In {\em ICCV}, pages 3392--3399, 2013.

\bibitem{fouhey2014unfolding}
David~Ford Fouhey, Abhinav Gupta, and Martial Hebert.
\newblock Unfolding an indoor origami world.
\newblock In {\em ECCV}, pages 687--702, 2014.

\bibitem{fu2018deep}
Huan Fu, Mingming Gong, Chaohui Wang, Kayhan Batmanghelich, and Dacheng Tao.
\newblock Deep ordinal regression network for monocular depth estimation.
\newblock In {\em CVPR}, pages 2002--2011, 2018.

\bibitem{Girshick2015Fast}
Ross Girshick.
\newblock Fast {R-CNN}.
\newblock In {\em ICCV}, pages 1440--1448, 2015.

\bibitem{Gupta2014Learning}
Saurabh Gupta, Ross Girshick, Pablo Arbel¨¢ez, and Jitendra Malik.
\newblock Learning rich features from rgb-d images for object detection and
  segmentation.
\newblock In {\em ECCV}, volume 8695, pages 345--360, 2014.

\bibitem{He2017Mask}
Kaiming He, Georgia Gkioxari, Piotr Doll¨¢r, and Ross Girshick.
\newblock Mask {R-CNN}.
\newblock {\em ICCV}.

\bibitem{he2013guided}
Kaiming He, Jian Sun, and Xiaoou Tang.
\newblock Guided image filtering.
\newblock {\em IEEE transactions on pattern analysis and machine intelligence},
  (6):1397--1409, 2013.

\bibitem{he2016deep}
Kaiming He, Xiangyu Zhang, Shaoqing Ren, and Jian Sun.
\newblock Deep residual learning for image recognition.
\newblock In {\em CVPR}, pages 770--778, 2016.

\bibitem{he2017std2p}
Yang He, Wei-Chen Chiu, Margret Keuper, Mario Fritz, and SI Campus.
\newblock Std2p: Rgbd semantic segmentation using spatio-temporal data-driven
  pooling.
\newblock In {\em CVPR}, pages 7158--7167, 2017.

\bibitem{He2016STD2P}
Yang He, Wei-Chen Chiu, Margret Keuper, Mario Fritz, and SI Campus.
\newblock Std2p: Rgbd semantic segmentation using spatio-temporal data-driven
  pooling.
\newblock pages 7158--7167, 2017.

\bibitem{Ke_2018_ECCV}
Tsung-Wei Ke, Jyh-Jing Hwang, Ziwei Liu, and Stella~X. Yu.
\newblock Adaptive affinity fields for semantic segmentation.
\newblock In {\em ECCV}, 2018.

\bibitem{kendall2015bayesian}
Alex Kendall, Vijay Badrinarayanan, , and Roberto Cipolla.
\newblock Bayesian segnet: Model uncertainty in deep convolutional
  encoder-decoder architectures for scene understanding.
\newblock {\em arXiv preprint arXiv:1511.02680}, 2015.

\bibitem{Kim2016Unified}
Seungryong Kim, Kihong Park, Kwanghoon Sohn, and Stephen Lin.
\newblock Unified depth prediction and intrinsic image decomposition from a
  single image via joint convolutional neural fields.
\newblock In {\em ECCV}, pages 143--159, 2016.

\bibitem{Kokkinos2017UberNet}
Iasonas Kokkinos.
\newblock Ubernet: Training a `universal' convolutional neural network for
  low-, mid-, and high-level vision using diverse datasets and limited memory.
\newblock In {\em CVPR}, pages 5454--5463, 2017.

\bibitem{kong2018pixel}
Shu Kong and Charless Fowlkes.
\newblock Pixel-wise attentional gating for parsimonious pixel labeling.
\newblock {\em arXiv preprint arXiv:1805.01556}, 2018.

\bibitem{krahenbuhl2011efficient}
Philipp Kr{\"a}henb{\"u}hl and Vladlen Koltun.
\newblock Efficient inference in fully connected crfs with gaussian edge
  potentials.
\newblock In {\em NIPS}, pages 109--117, 2011.

\bibitem{Laina2016Deeper}
Iro Laina, Christian Rupprecht, Vasileios Belagiannis, Federico Tombari, and
  Nassir Navab.
\newblock Deeper depth prediction with fully convolutional residual networks.
\newblock In {\em International Conference on 3D Vison}, pages 239--248, 2016.

\bibitem{levin2008closed}
Anat Levin, Dani Lischinski, and Yair Weiss.
\newblock A closed-form solution to natural image matting.
\newblock {\em IEEE transactions on pattern analysis and machine intelligence},
  30(2):228--242, 2008.

\bibitem{li2018monocular}
Bo Li, Yuchao Dai, and Mingyi He.
\newblock Monocular depth estimation with hierarchical fusion of dilated cnns
  and soft-weighted-sum inference.
\newblock {\em Pattern Recognition}, 2018.

\bibitem{li2015depth}
Bo Li, Chunhua Shen, Yuchao Dai, Anton van~den Hengel, and Mingyi He.
\newblock Depth and surface normal estimation from monocular images using
  regression on deep features and hierarchical crfs.
\newblock In {\em CVPR}, pages 1119--1127, 2015.

\bibitem{li2018deep}
Ruibo Li, Ke Xian, Chunhua Shen, Zhiguo Cao, Hao Lu, and Lingxiao Hang.
\newblock Deep attention-based classification network for robust depth
  prediction.
\newblock {\em arXiv preprint arXiv:1807.03959}, 2018.

\bibitem{Li2016LSTM}
Zhen Li, Yukang Gan, Xiaodan Liang, Yizhou Yu, Hui Cheng, and Liang Lin.
\newblock Lstm-cf: Unifying context modeling and fusion with lstms for rgb-d
  scene labeling.
\newblock In {\em ECCV}, pages 541--557, 2016.

\bibitem{Lin2016RefineNet}
Guosheng Lin, Anton Milan, Chunhua Shen, and Ian Reid.
\newblock Refinenet: Multi-path refinement networks for high-resolution
  semantic segmentation.
\newblock In {\em CVPR}, volume~1, pages 5168--5177, 2017.

\bibitem{Lin2016Efficient}
Guosheng Lin, Chunhua Shen, Anton Van~Den Hengel, and Ian Reid.
\newblock Efficient piecewise training of deep structured models for semantic
  segmentation.
\newblock In {\em CVPR}, pages 3194--3203, 2016.

\bibitem{liu2015Learning}
Fayao Liu, Chunhua Shen, Guosheng Lin, and Ian Reid.
\newblock Learning depth from single monocular images using deep convolutional
  neural fields.
\newblock {\em IEEE Transactions on Pattern Analysis and Machine Intelligence},
  38(10):2024--2039, 2016.

\bibitem{liu2016learning}
Risheng Liu, Guangyu Zhong, Junjie Cao, Zhouchen Lin, Shiguang Shan, and
  Zhongxuan Luo.
\newblock Learning to diffuse: A new perspective to design pdes for visual
  analysis.
\newblock {\em IEEE transactions on pattern analysis and machine intelligence},
  38(12):2457--2471, 2016.

\bibitem{liu2017learning}
Sifei Liu, Shalini De~Mello, Jinwei Gu, Guangyu Zhong, Ming-Hsuan Yang, and Jan
  Kautz.
\newblock Learning affinity via spatial propagation networks.
\newblock In {\em NIPS}, pages 1520--1530, 2017.

\bibitem{Long2017Fully}
Jonathan Long, Evan Shelhamer, and Trevor Darrell.
\newblock Fully convolutional networks for semantic segmentation.
\newblock {\em IEEE Transactions on Pattern Analysis and Machine Intelligence},
  39(4):640--651, 2017.

\bibitem{Misra2016Cross}
Ishan Misra, Abhinav Shrivastava, Abhinav Gupta, and Martial Hebert.
\newblock Cross-stitch networks for multi-task learning.
\newblock In {\em CVPR}, pages 3994--4003, 2016.

\bibitem{Kundu_2018_CVPR}
Jogendra Nath~Kundu, Phani Krishna~Uppala, Anuj Pahuja, and R. Venkatesh~Babu.
\newblock Adadepth: Unsupervised content congruent adaptation for depth
  estimation.
\newblock In {\em CVPR}, 2018.

\bibitem{Noh2015Learning}
Hyeonwoo Noh, Seunghoon Hong, and Bohyung Han.
\newblock Learning deconvolution network for semantic segmentation.
\newblock In {\em ICCV}, pages 1520--1528, 2015.

\bibitem{paszke2017automatic}
Adam Paszke, Sam Gross, Soumith Chintala, Gregory Chanan, Edward Yang, Zachary
  DeVito, Zeming Lin, Alban Desmaison, Luca Antiga, and Adam Lerer.
\newblock Automatic differentiation in pytorch.
\newblock 2017.

\bibitem{qi20173d}
Xiaojuan Qi, Renjie Liao, Jiaya Jia, Sanja Fidler, and Raquel Urtasun.
\newblock 3d graph neural networks for rgbd semantic segmentation.
\newblock In {\em CVPR}, pages 5199--5208, 2017.

\bibitem{qi2018geonet}
Xiaojuan Qi, Renjie Liao, Zhengzhe Liu, Raquel Urtasun, and Jiaya Jia.
\newblock Geonet: Geometric neural network for joint depth and surface normal
  estimation.
\newblock In {\em CVPR}, pages 283--291, 2018.

\bibitem{Roy2016CVPR}
Anirban Roy and Sinisa Todorovic.
\newblock Monocular depth estimation using neural regression forest.
\newblock In {\em CVPR}, pages 5506--5514, 2016.

\bibitem{SJ2017RDF}
Park Seong-Jin, Hong Ki-Sang, and Lee Seungyong.
\newblock Rdfnet: Rgb-d multi-level residual feature fusion for indoor semantic
  segmentation.
\newblock In {\em ICCV}, pages 4990--4999, 2017.

\bibitem{silberman2012indoor}
Nathan Silberman, Derek Hoiem, Pushmeet Kohli, and Rob Fergus.
\newblock Indoor segmentation and support inference from {RGBD} images.
\newblock In {\em ECCV}, pages 746--760, 2012.

\bibitem{simonyan2014very}
Karen Simonyan and Andrew Zisserman.
\newblock Very deep convolutional networks for large-scale image recognition.
\newblock {\em arXiv preprint arXiv:1409.1556}, 2014.

\bibitem{Song2015SUN}
S. Song, S.~P. Lichtenberg, and J. Xiao.
\newblock Sun {RGB-D}: A {RGB-D} scene understanding benchmark suite.
\newblock In {\em CVPR}, pages 567--576, 2015.

\bibitem{tateno2017cnn}
Keisuke Tateno, Federico Tombari, Iro Laina, and Nassir Navab.
\newblock Cnn-slam: Real-time dense monocular slam with learned depth
  prediction.
\newblock In {\em CVPR}, volume~2, 2017.

\bibitem{uhrig2017sparsity}
Jonas Uhrig, Nick Schneider, Lukas Schneider, Uwe Franke, Thomas Brox, and
  Andreas Geiger.
\newblock Sparsity invariant cnns.
\newblock In {\em International Conference on 3D Vision}, pages 11--20, 2017.

\bibitem{Wang2015Towards}
Peng Wang, Xiaohui Shen, Zhe Lin, and Scott Cohen.
\newblock Towards unified depth and semantic prediction from a single image.
\newblock In {\em CVPR}, pages 2800--2809, 2015.

\bibitem{wang2016surge}
Peng Wang, Xiaohui Shen, Bryan Russell, Scott Cohen, Brian Price, and Alan~L
  Yuille.
\newblock Surge: Surface regularized geometry estimation from a single image.
\newblock In {\em NIPS}, pages 172--180, 2016.

\bibitem{wang2015designing}
Xiaolong Wang, David Fouhey, and Abhinav Gupta.
\newblock Designing deep networks for surface normal estimation.
\newblock In {\em CVPR}, pages 539--547, 2015.

\bibitem{wang2017non}
Xiaolong Wang, Ross Girshick, Abhinav Gupta, and Kaiming He.
\newblock Non-local neural networks.
\newblock In {\em CVPR}, volume~10, 2018.

\bibitem{xu2018pad}
Dan Xu, Wanli Ouyang, Xiaogang Wang, and Nicu Sebe.
\newblock Pad-net: Multi-tasks guided prediction-and-distillation network for
  simultaneous depth estimation and scene parsing.
\newblock In {\em CVPR}, 2018.

\bibitem{Xu2017CVPR}
Dan Xu, Elisa Ricci, Wanli Ouyang, Xiaogang Wang, and Nicu Sebe.
\newblock Multi-scale continuous {CRF}s as sequential deep networks for
  monocular depth estimation.
\newblock In {\em CVPR}, volume~1, pages 161--169, 2017.

\bibitem{xu2018structured}
Dan Xu, Wei Wang, Hao Tang, Hong Liu, Nicu Sebe, and Elisa Ricci.
\newblock Structured attention guided convolutional neural fields for monocular
  depth estimation.
\newblock In {\em CVPR}, pages 3917--3925, 2018.

\bibitem{zeisl2014discriminatively}
Bernhard Zeisl, Marc Pollefeys, et~al.
\newblock Discriminatively trained dense surface normal estimation.
\newblock In {\em ECCV}, pages 468--484, 2014.

\bibitem{zhang2018joint}
Zhenyu Zhang, Zhen Cui, Chunyan Xu, Zequn Jie, Xiang Li, and Jian Yang.
\newblock Joint task-recursive learning for semantic segmentation and depth
  estimation.
\newblock In {\em ECCV}, pages 235--251, 2018.

\bibitem{zhou2017unsupervised}
Tinghui Zhou, Matthew Brown, Noah Snavely, and David~G Lowe.
\newblock Unsupervised learning of depth and ego-motion from video.
\newblock In {\em CVPR}, volume~2, page~7, 2017.

\end{thebibliography}
}

\end{document}